\documentclass{svproc}
\usepackage{times}
\usepackage{multicol}
\usepackage{graphicx}
\usepackage{subfigure}
\usepackage{epsfig}
\usepackage{amsmath,amssymb,bm}
\usepackage{algorithmic}
\usepackage{url}
\usepackage{color}
\usepackage[ruled]{algorithm2e}
\usepackage{ntheorem}
\usepackage{float}
\usepackage{physics}
\usepackage{cite}
\usepackage{todonotes}
\usepackage{mathtools}
\usepackage[normalem]{ulem}
\usepackage{svg}
\usepackage{soul}
\usepackage[bottom]{footmisc}
\usepackage{hyperref}
\usepackage{cleveref}

\newcommand{\etal}{\textit{et al. }}

\usepackage{url}

\begin{document}
\mainmatter              % start of a contribution
\title{\textit{\small Accepted to Distributed Autonomous Robotic Systems 2024 (DARS) 2024}\\[1em]
Challenges Faced by Large Language Models in Solving Multi-Agent Flocking}
\titlerunning{Challenges Faced By Large Language Models In Solving Multi-Agent Flocking}  % abbreviated title (for running head)
%                                     also used for the TOC unless
%                                     \toctitle is used
%
\author{Peihan Li \and Vishnu Menon \and
Bhavanaraj Gudiguntla \and Daniel Ting \and
Lifeng Zhou$^\dagger$}
\authorrunning{Peihan Li et al.} % abbreviated author list (for running head)
%
%%%% list of authors for the TOC (use if author list has to be modified)
% \tocauthor{Ivar Ekeland, Roger Temam, Jeffrey Dean, David Grove,
% Craig Chambers, Kim B. Bruce, and Elisa Bertino}
%
\institute{Drexel University, Philadelphia, PA 19104, USA\\
\email{\{pl525, vvm33, bg653, dct55, lz457\}@drexel.edu},
\\ $^\dagger$ Corresponding author
% YouTube link:
% \texttt{http://users/\homedir iekeland/web/welcome.html}
% \and
% Universit\'{e} de Paris-Sud,
% Laboratoire d'Analyse Num\'{e}rique, B\^{a}timent 425,\\
% F-91405 Orsay Cedex, France
}

\maketitle              % typeset the title of the contribution

\begin{abstract}

%\LZ{flocking, why flocking, LLMs, why we use LLMs for flocking, what we try, results:simulations and experiments. }

\vspace{-3mm}
Flocking is a behavior where multiple agents in a system attempt to stay close to each other while avoiding collision and maintaining a desired formation. This is observed in the natural world and has applications in robotics, including search and rescue, wild animal tracking, and perimeter surveillance. Recently, large language models (LLMs) have displayed an impressive ability to solve various collaboration tasks as individual decision-makers. Solving multi-agent flocking with LLMs would demonstrate their usefulness in situations requiring spatial and decentralized decision-making. Yet, when LLM-powered agents are tasked with implementing multi-agent flocking, they fall short of the desired behavior. After extensive testing, we find that agents with LLMs as individual decision-makers typically opt to converge on the average of their initial positions or diverge from each other. After breaking the problem down, we discover that LLMs cannot understand maintaining a shape or keeping a distance in a meaningful way. Solving multi-agent flocking with LLMs would enhance their ability to understand collaborative spatial reasoning and lay a foundation for addressing more complex multi-agent tasks. This paper discusses the challenges LLMs face in multi-agent flocking and suggests areas for future improvement and research.

% The abstract should summarize the contents of the paper
% using at least 70 and at most 150 words. It will be set in 9-point
% font size and be inset 1.0 cm from the right and left margins.
% There will be two blank lines before and after the Abstract. \dots
% We would like to encourage you to list your keywords within
% the abstract section using the \keywords{...} command.
\end{abstract}
\section{Introduction}
Large language models (LLMs) have recently emerged as a hot topic in academic and research communities. As an artificial intelligence (AI) technique trained on a large number of diverse datasets, LLMs are showing reasoning abilities that exceed the traditional language tasks like text generation and translation and acting as powerful problem solvers in various domains. This reasoning ability helps LLMs solve mathematical problems~\cite{wu2024mathchat}, provide medical advice based on the user description~\cite{gao2023large}, and generate plans to accomplish tasks~\cite{valmeekam2023planbench}. 

%All those applications often require an efficient flocking behavior to provide a better group formation. 

Motivated by the reasoning and problem-solving ability observed with the recent development of LLMs, some researchers are exploring applying LLMs to solve multi-agent problems. Multi-agent systems play an important role in accomplishing real-world tasks, including natural disaster search and rescue~\cite{rahman2022adversar}, wild animal tracking, and perimeter surveillance and patrol~\cite{banerjee2024decentralized}. Explorations in using LLMs for decentralized planning~\cite{chen2024scalable} and consensus solving~\cite{chen2023multi} have shown fruitful results in task allocation and decision-making. However, multi-agent flocking is rarely explored with the application of LLMs. We aim to rectify this research gap and examine the potential for applying LLMs to the wider task of multi-agent flocking. In a seminal work of flocking~\cite{reynolds1987flocks}, Reynolds establishes three rules to define flocking behavior. They are -
Collision Avoidance: avoid collisions with nearby flockmates, Velocity Matching: attempt to match velocity with nearby flockmates, and Flock Centering: attempt to stay close to nearby flockmates.
These rules, referred to as Boids (short for bird-oids) rules for flocking, are how we define successful flocking. The flocking behavior can be commonly observed in flocks of birds, bees, and schools of fish, where each agent in the group is constantly making its own decisions based on the behavior of its neighboring agents. This decentralized decision-making nature makes the multi-agent flocking a complex problem to solve. Previous works for multi-agent flocking provide fixed~\cite{tanner2003stable} and dynamic~\cite{olfati2006flocking} interactions.

% core idea from the multi-agent path finding paper
% https://arxiv.org/pdf/2401.03630.pdf
% \vspace{-2mm}
Trained on large datasets, LLMs contain a variety of information and do not require fine-tuning to solve many problems. Many recent works like ~\cite{li2023camel}, ~\cite{chen2024scalable}, ~\cite{tan2023language}, ~\cite{hong2023metagpt} and~\cite{chen2023multi} have been successful in applying innovative solutions towards multi-agent systems and robotics. Just as these problems have real-world applications, solving flocking has real applications to environmental monitoring~\cite{sung2023survey}, target tracking~\cite{edwards2022stochastic}, and surveillance~\cite{rahmani2019flocking}. Additionally, solving flocking with LLMs might enable them to solve other complication problems related to logic, spatial, and collaborative reasoning-- increasing their general intelligence~\cite{chen2023multi}.

\noindent \textbf{Contributions}. We introduce a decentralized decision-making model using LLMs for multi-agent flocking. Each agent is equipped with its own LLM, simulating a brain that independently processes prompts consisting of the agent's role, game rules, and dynamic round descriptions. These prompts guide the agents to navigate, coordinate, and ultimately form a specified flock shape, adhering to predefined rules and considering the agents' positions.
% \LZ{and velocities}
This decentralized approach contrasts with centralized models by empowering each agent with autonomy and enhancing problem-solving capabilities in complex, collaborative tasks.

% Results - what kind of experiments we tried
Experiments are conducted to evaluate the capability of LLMs in multi-agent flocking scenarios, involving formations such as circles, triangles, and lines with agents maintaining specific distances. We perform tests with GPT-3.5-Turbo and GPT-4-Turbo to form flocks with varying agent numbers and configurations. However, GPT-4-Turbo has a high failure rate due to its disregard for the specified output format. Therefore, we focus on the tests using GPT-3.5-Turbo. Our investigations reveal that LLMs like GPT-3.5-Turbo face significant challenges in solving the multi-agent flocking problem, chiefly due to their lack of spatial and collaborative reasoning. These models struggle with the fundamental aspects of flocking, such as coordinating movements, maintaining specific formations, and ensuring proper distance between agents, with their only success being convergence towards a singular point. This limitation underscores the urgent need for enhancing LLMs' capabilities in spatial awareness and reasoning to address complex, real-world problems effectively.
% The experiments highlight challenges in agents achieving desired formations, often resulting in gathering or failing to maintain required distances, indicating limitations in the LLMs' spatial reasoning capabilities.
% . The experiments highlighted challenges in agents achieving desired formations, often resulting in gathering or failing to maintain required distances, indicating limitations in the LLMs' spatial reasoning capabilities.

% \LZ{this is good, but needs more description since this is the most important part of the intro: Method: how we use LLMs for flocking, what kind of prompt we use, we should also emphasize decentralized LLMs here, each agent has a LLM. it is not a centralized LLM that makes decisions for all agents. Results: what kind of experiments we try. Conclusion: fail, why it fails.}

% ADD APPROACH HERE
% ADD SIGNIFICANCE(?) HERE
% CONTRIBUTIONS OR KEY FINDINGS

% - what is the flocking, why flocking is important (motivation)- copy paste 

% - why use LLMs to solve flocking (refer to the LLMs for consensus paper~\cite{chen2023multi}.) 
% what is LLMs? a short description. 

% - approach: how we use LLMs for flocking 

% - contributions or key findings: 
% results, situations, potential reasons
% 1, 2, 3 

\section{Related Work}
\vspace{-3mm}

\subsection{Multi-agent Flocking}
Multi-agent flocking refers to the coordinated movement of a group of independent agents in a way that resembles the natural flocking behavior of birds. This topic is studied to understand how groups of agents can work together without a leader to achieve the desired formation. The paper~\cite{reynolds1987flocks} highlights the three flocking rules of Reynolds namely: Flock Centering, Collision Avoidance, and Velocity Matching. Based on these rules, Olfati-Saber has demonstrated three different flocking algorithms~\cite{olfati2006flocking} for $\alpha$ agents. An $\alpha$ agent's main objective is to create an $\alpha$-lattice with its neighboring $\alpha$-agents. The first Algorithm generates spatial order in flocks, incorporating all three rules of Reynolds in one equation, leading to regular fragmentation. The second Algorithm leads to flocking in free space, with an added consideration for $\gamma$ agents for group objectives. The third Algorithm combines the first two algorithms and obstacle avoidance capabilities provided by virtual agents known as $\beta$ agents. Recent work in multi-agent flocking includes studying the two collective behaviors like swarming and flocking~\cite{sar2023flocking}. Flocking involves agents aligning their motion, whereas swarming involves them gathering together to coordinate their spatial arrangement. It is observed that inter-particle avoidance is a necessary condition for flocking. Li \etal~\cite{LI20089344} proposed the artificial potential-based approach for the migration and trajectory tracking of multi-agent systems. Zhu \etal~\cite{9169650} utilized deep reinforcement learning to learn multi-agent flocking control in complex environments with dynamic obstacles. Multi-robot flocking control using Multi-Agent Twin Delayed Deep Deterministic Policy Gradient (MATD3) has been studied to solve the issue of overestimation bias~\cite{9975963}. Chen \etal~\cite{9187614} has discussed how dividing the flocking group into subgroups can help to address the problem of maintaining social distancing among them and allow for better coordination. This study~\cite{10484985} proposes a discrete-time flocking control model for multi-robot systems with random link failures, ensuring practical implementation and achieving flocking in expectation with improved stability conditions. 
\vspace{-3mm}

\subsection{LLMs for Multi-agent Systems}
 LLM is an AI program that can recognize natural language and solve problems. LLMs can act as individual decision-makers and are used for decentralized applications. Recent emerging research has been focused on exploring the applications of LLMs in multi-agent systems. One such work~\cite{du2023improving} focused on improving the LLM responses. A new approach has been presented where multiple language model instances collaborate to debate their reasoning and responses for various tasks. Multiple LLMs are also used in the LCTGen~\cite{tan2023language} model to generate dynamic traffic scenarios. Recent work~\cite{li2023camel} shows that using multiple LLM agents and allowing conversation between them to create a feedback loop is useful for solving complex tasks like instruction-following cooperation. Agashe \etal~\cite{agashe2023evaluating} evaluates the reasoning ability of LLMs to achieve multi-agent coordination in three different environments. It is proved that LLMs can outperform MARL methods in one of those layouts. LLMs have also been used to create dialogue games~\cite{schlangen2023dialogue} to measure the ability to understand languages. Another work on LLMs for multi-agent systems is CLIPSwarm~\cite{pueyo2024clipswarm} a new algorithm used to automate the modeling of swarm agents formation. Chen \etal~\cite{chen2023multi} highlighted how when multiple LLMs are used to complete a common task, they might initially have different solutions. This work focuses on how LLM-driven consensus-seeking can be applied to the multi-agent task assignment process. This process showcases the capability of LLM-driven agents to accomplish autonomous planning seamlessly for collaborative tasks among multiple robots. The primary findings of this work are mainly - consensus strategy (agents tend to have collaborative behavior), impact of personality (stubborn and suggestible), impact of agent number (having more agents reduces the randomness of the system), and impact of topology (fully connected networks are most efficient). Chen \etal~\cite{chen2024scalable} demonstrates how LLMs can be effective multi-agent task planners. LLMs are also used in~\cite{hong2023metagpt} MetaGPT for including human workflows for performing multi-agent collaboration tasks. 
 \vspace{-3mm}

\subsection{LLMs for Reasoning}
The exploration of enhancing the reasoning abilities of LLMs has been a focal point of recent scholarly efforts. The introduction of Chain-Of-Thought (CoT) prompting \cite{wei2022chain}, marks a significant advancement in bolstering LLMs' reasoning capabilities. Unlike traditional fine-tuning methods, CoT prompting encourages models to mimic human-like reasoning processes, leading to superior performance. However, other research suggests that the current general planning capabilities of LLMs are still rather lackluster\cite{valmeekam2023planbench}. Furthermore, current research has been focused on developing innovative techniques to improve the reasoning capabilities of LLMs despite their flaws. Some examples include prepending a Task-Agnostic Prefix Prompt \cite{ye2023investigating}, selecting an optimal base model for reasoning prior for reasoning \cite{zhao2023automatic}, and introducing incremental hints to LLMs \cite{zheng2023progressive}. In another novel methodology, Meng \etal~\cite{meng2024divide} explores the strategic division of a question dataset based on the inferred confidence scores of an LLM, followed by a detailed analysis of question complexity through diverse methods. Lastly, Lu \etal~\cite{lu2023chameleon} unveil the Chameleon framework, a `plug and play' approach that augments LLM functionalities with additional modules, including web searches, image captioning, text detection, and reasoning sequencing. This methodology not only broadens the operational scope of LLMs but also parallels the enhanced capabilities observed in the latest iteration of ChatGPT-4.

\vspace{-5mm}
\section{LLMs for Multi-agent Flocking}
\vspace{-2mm}

\subsection{Preliminaries}

\subsubsection{Multi-agent flocking}

%Multi-agent flocking is a well-studied problem. Reynolds\cite{reynolds1987flocks} proposed the first approach to simulate the flocking behavior of multiple agents for computer animation purposes in 1986. Later, an algorithm for mobile agents with a double integrator dynamic model was proposed by Tanner \etal~\cite{tanner2003stable}. Then, Olfati-Saber proposed a decentralized approach to the multi-agent flocking problem with similar double integrator dynamics~\cite{olfati2006flocking}. 

Olfati-Saber formulated the control laws consist a combination of attractive/repulsive, alignment, and leading forces~\cite{olfati2006flocking}, which tackled the collision avoidance and cohesion of the group, the common heading direction for an aggregate motion, as well as the potential goal location for the flock to reach. 
% We take Olfati-Saber's algorithm as the baseline policy to evaluate the performance of LLMs in solving the multi-agent flocking problems. 

For a group of agents $N$, the double integrator dynamics model for agent $i\in N$ is described by: 
\begin{subequations} \label{eq:agent_motion}
    \begin{align}
        \dot{q_i} &= v_i    \\
        \dot{v_i} &= u_i
    \end{align}
\end{subequations}
where $q_i = [q_i, q_j]$ denotes its position, $v_i = [\dot{v_i}, \dot{v_j}]$ denotes its velocity, and $u_i = [u_{x_i}, u_{y_i}]$ represents its control input (acceleration). The flocking algorithm introduced in~\cite{olfati2006flocking} is $u_i = u_i^\alpha$ with 
\begin{equation}    \label{eq:algo_1}
    u_i^\alpha = \underbrace{\sum_{j\in{N_i}} \phi_{\alpha} (\| q_j - q_i \|_{\sigma}) \mathbf{n}_{ij}}_\text{gradient-based term} + \underbrace{\sum_{j\in{N_i}} a_{ij} (q) (v_j - v_i)}_\text{consensus term}
\end{equation}
where:
\begin{itemize}
\vspace{-3mm}
    \item the $\alpha$-term controls the distance between agents to avoid collision by providing attraction forces based on the distance between agent $i$ and $j$, 
    \item $N_i = \{ j \in \nu : \| q_j - q_i \| < r \}$ is the set of spatial neighbors of agent $i$ within interaction range $r$, 
    \item $\mathbf{n}_{ij} = \sigma_{\epsilon} (q_j - q_i) = \frac{(q_j - q_i)}{\sqrt{1 + \epsilon \| q_j - q_i \| ^2}}$ is a vector along the line joining $q_i$ to $q_j$, 
    \item $\epsilon \in (0,1)$ is a fixed $\sigma$-norm parameter, 
    \item $\phi_{\alpha}$  is the action function to construct a smooth pairwise potential with finite cut-off, 
    \item and $a_{ij}(q)$ is the element used to define spatial adjacency matrix $A(q)$.
\end{itemize}

\subsubsection{LLMs for consensus}

%LLMs for consensus - 
%how they use LLMs for consensus.
In their foundational experiment, Chen \etal \cite{chen2023multi} explored the concept of consensus-seeking among multiple LLM agents, a task inspired by the broader field of robotics-based consensus. However, the scope of their study is intentionally narrowed, focusing on the behavior of agents within a 1D space. In this setup, n (\(n \geq 2\)) agents exist, each with a value between 0-100 representing some abstract quantity. In this example, the value is described as each agent's position. The goal is for all agents to converge at a singular point with each agent deciding their next position based on their current position and other agents' positions. The environment updates after each discrete time step known as a round. The authors use Chain-Of-Thought reasoning to enhance the performance of their zero-shot prompts. No specific strategy is mentioned in the prompt itself; the goal is for the agents to figure out their strategy to reach a consensus. They first give an initial prompt to each agent, describing the original environment parameters and response format: 
\begin{quote}
\colorbox{yellow!10}{\parbox{11.3cm}
{``You are an agent moving in a one-dimensional space."}}
\end{quote}
\begin{quote}\colorbox{yellow!10}{\parbox{11.3cm}
{
``Another agent is present in the space,
and you need to gather. Your position is:
39 and the other agent's position is: [87].
You need to choose a position to move to
in order to gather, and briefly explain the
reasoning behind your decision. Strictly
follow the 'Reasoning: . . . , Position: . . .'
format to provide your answer. In the
'Reasoning' section, it is your thought
process, while the 'position' section is only
the location you wish to move to in this
round, without any further explanation
needed"}}
\end{quote}
The LLM then responds by calculating a new position with a given reason:
\begin{quote}\colorbox{cyan!10}{\parbox{11.3cm}
{
``Reasoning: In order to gather with the
other agent, I need to move towards their
position. Since the other agent is at
position 87 and I am at position 39, I need
to move towards the right direction to
reach their position. Position: 40``}}
\end{quote}
Afterwards, the system dynamically prompts each agent with a repeating prompt until the maximum set rounds have passed:
\begin{quote}\colorbox{yellow!10}{\parbox{11.3cm}
{
``You have moved to 40.0, and the latest
position of another agent is: [60.0], please
choose the position you want to move to
next."}}
\end{quote}
Chen \etal~\cite{chen2023multi} also allocate personalities to different agents through prompting. The stubborn personality would make the agent not move and the suggestible personality would make the agent move to another agent's exact position:
\begin{quote}\colorbox{yellow!10}{\parbox{11.3cm}
{
``You are an extremely stubborn person, prefer
to remain stationary."}}
\end{quote}
\begin{quote}\colorbox{yellow!10}{\parbox{11.3cm}
{
``You are an extremely suggestible person,
prefer to move to someone else's position."}}
\end{quote}
This changes the reasoning strategies of the agents, resulting in oscillations or multiple convergence points. They also explore network topologies by only giving some agents information about other specific agents. For example, in a leader-follower structure, the leader agent does not know the existence of the other agents and the other agents only know the existence of the leader agent. The environment is later expanded to two dimensions and could potentially work in higher dimensions. Chen \etal~\cite{chen2023multi} acknowledged certain limitations in their research, including questions about the scalability of this consensus-seeking behavior to other tasks, its effectiveness across different LLM architectures, and concerns over the infrequent updates of the planner.

% \vspace{-3mm}
\subsection{LLMs for flocking}
%gpt to do flocking, prompt design,

%\LZ{why we don't use velocity? give some explanations, otherwise the reviewer will ask}

%\BG{We do not use velocity as it might make it difficult to interpret the decision-making process. LLMs are stateless models and velocity is inherently a state concept. I am not sure of any other reasons.}

We use LLMs as decentralized decision-makers for agents to achieve flocking behaviors. For each agent, a unique prompt is created and called to an LLM. The LLM responds to that prompt, representing the brain of a single agent instead of controlling all agents in a centralized manner. 
% Technically speaking, LLMs are inherently stateless so whether the LLM instances are the same for different agents doesn't matter. Instead, 
The proper context must be provided to an LLM to perform actions based on an agent's previous history, simulating a separate decentralized decision-maker. The prompt explained in detail below has 3 parts: agent role, game description, and round description. The agent role and game description are given to each agent's LLM in the first round of decision-making. Each agent's LLM decides where it should move, based on the rules given in the game description. Each agent's LLM gives the output containing the next position and reasoning. This updated information is given as the round description in each round after the first round. This is continued until the flock is achieved for the number of rounds specified. 

   \begin{quote}
    \colorbox{cyan!10}{\parbox{11.3cm} {agent\_role = `` You are a agent navigating a two-dimensional space."

        }}
   \end{quote}
   
      \begin{quote}
    \colorbox{cyan!10}{\parbox{11.3cm} { game\_description = 
    % \\ \hspace{0.5cm} 
        `` There are other agents in the space, and you must coordinate with each other to form a flock of a specified shape. Keep in mind Boids flocking rules. Your position is: [{}]. The positions of the other agents (in the format [[x, y], [x, y]...]) are: [{}]. The maximum velocity is [{}] units per round. The flock shape is a [{}]. You must avoid getting closer than [{}] units to any peers, otherwise, you may collide. Remember to consider the positions and velocity of other agents and consider how they might behave. You need to choose a position to move to in order to form a flock, and briefly explain the reasoning behind your decision."
        
    }}
   \end{quote}
   
      \begin{quote}
    \colorbox{yellow!10}{\parbox{11.3cm} {
    round\_description = ``You have now moved to: [{}]. The new positions of the other agents are: [{}]. Consider how well your strategy worked last round, keeping in mind your maximum velocity, Please select a new position to move to."
        }}
   \end{quote}

      \begin{quote}
    \colorbox{yellow!10}{\parbox{11.3cm} {
    
    output\_format = 
    ``Strictly follow the `Reasoning:..., Position: [x, y]' format to provide your answer. x and y must both be floating point numbers truncated to two decimal places. Briefly provide your thought process in the reasoning section while keeping the position section ONLY for the position you wish to move to this iteration, without any further explanation. Do not write ANYTHING ELSE in the position section."

    }}
   \end{quote}

%%%%%%%%%%% MAE plot %%%%%%%%%%%%%%%%
\begin{figure*}[th!]
\vspace{-5.5mm}
    \centering
    \subfigure[Five active agents form circle with desired distance of 5 units.]{\includegraphics[width=0.45\columnwidth]{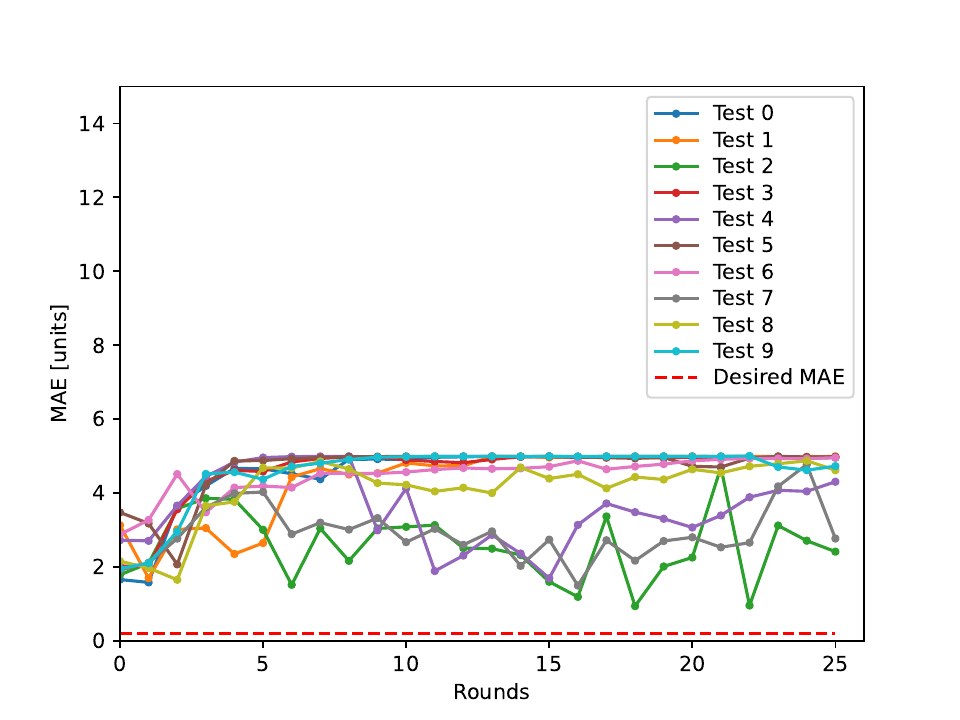} \label{fig:flock-circle-err}} 
    \subfigure[Three active agents form triangle with desired distance of 5 units.]{\includegraphics[width=0.45\columnwidth]{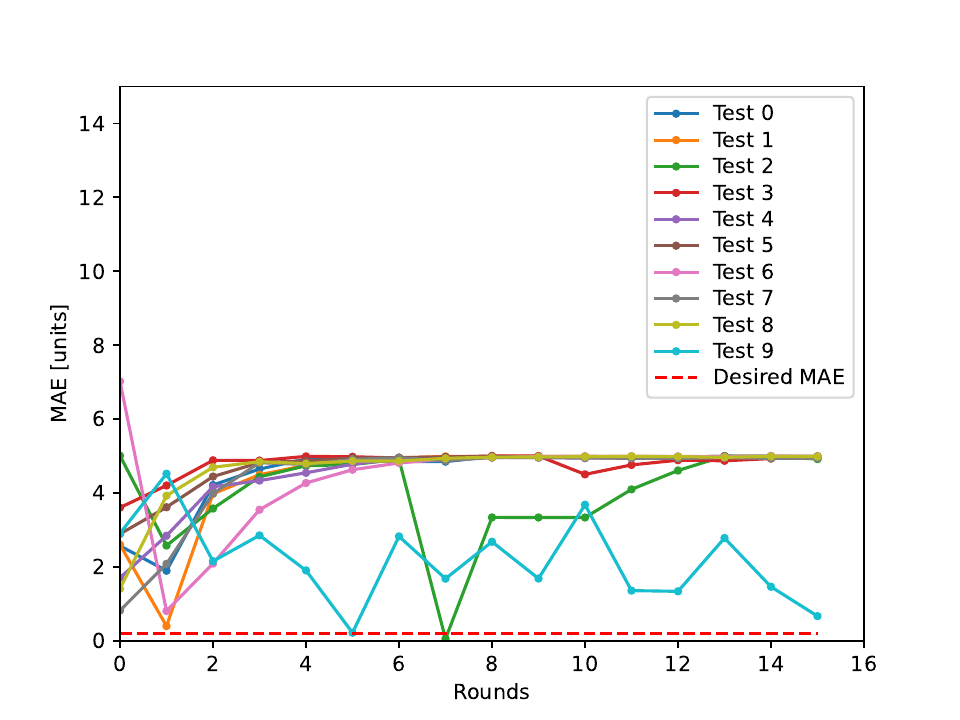} \label{fig:tri-0-err}} \\
    \vspace{-4mm}
    \subfigure[Two active agents keep desired distance of 10 units.]{\includegraphics[width=0.45\columnwidth]{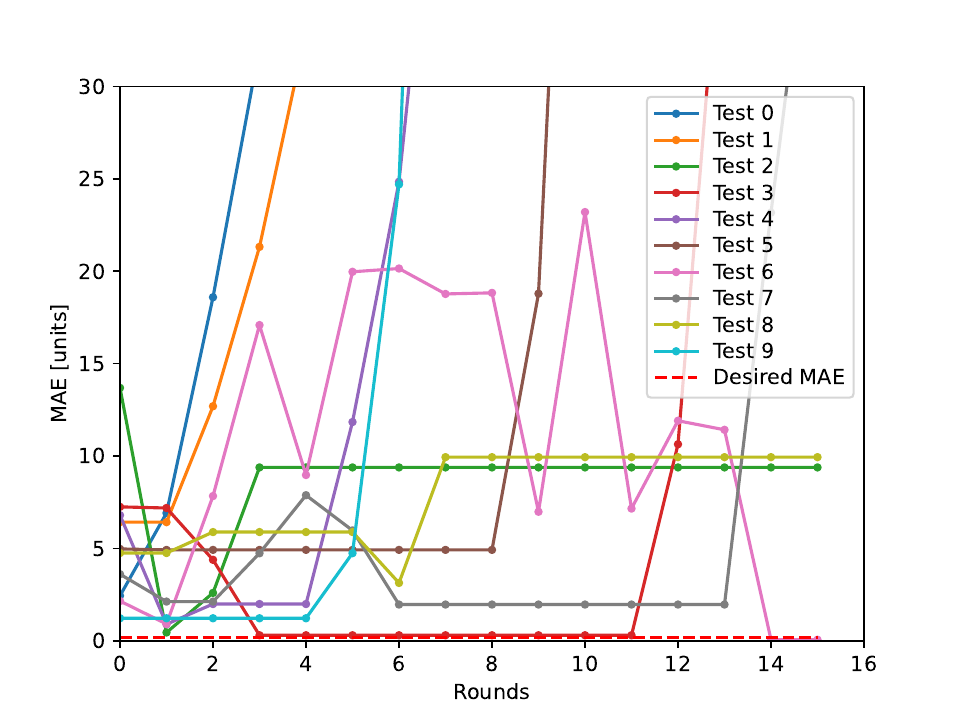} \label{fig:dist-0-err}} 
    \subfigure[One active agent, one stationary agent keep desired distance of 10 units.]{\includegraphics[width=0.45\columnwidth]{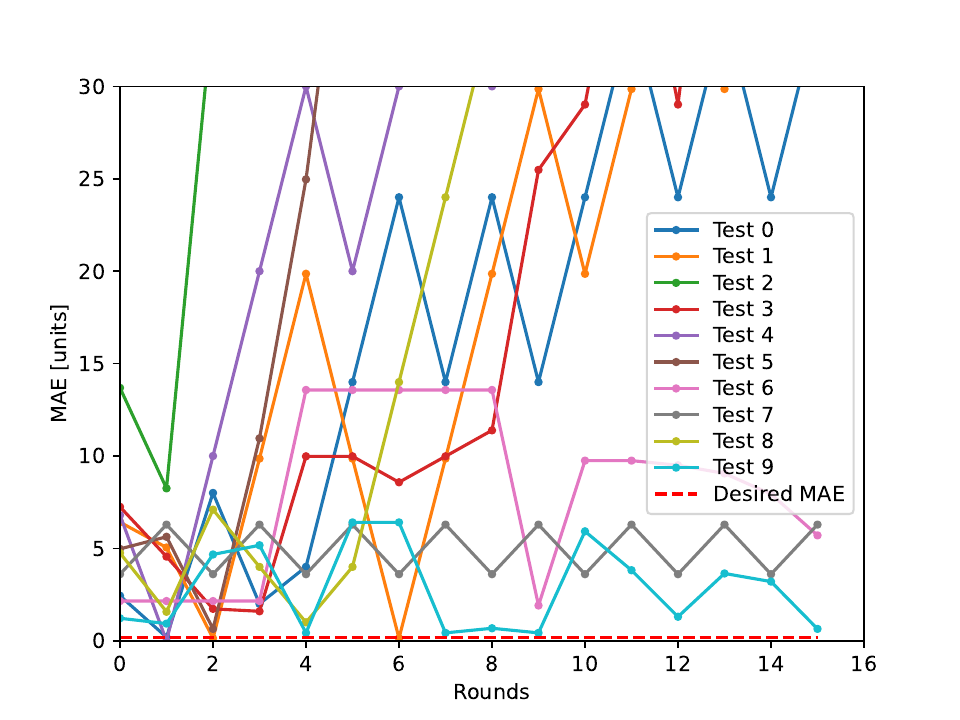} \label{fig:dist-1-err}}
    \vspace{-3mm}
    \caption{MAE of ten tests for different numbers of agents and different flock formations. The red dashed line shows the desired MAE (0.2 margin).}
    
\end{figure*}

Here, the ``agent\_role" is the role of each agent in flocking. The ``game\_description" explains the rules and goals of the game. The game aims to form a flock with all agents by keeping the Boid flocking rules in mind. The hyperparameters specified are the positions of the agent and its neighboring agents, the flock shape, the safe distance, and the maximum velocity (maximum units the agent can move). The agent role and game description are provided to the LLM in the first round. The ``round\_description", provided in the subsequent rounds, contains the updated positions and reminds the LLM of the maximum velocity. The ``output\_format", `Reasoning: ..., Position: [x,y]', describes the expected output structure. Note that, here we only provide an upper bound for the velocity but do not directly control it. The reason is twofold. Firstly, adding velocity control along with position control gives another layer of complexity for the LLMs to reason and understand, making it difficult for them to interpret the decision-making process. Secondly, we want to provide minimal information to the LLMs to rely more on them to reason and make decisions.
\vspace{-3mm}
% metrics for flocking behavior. 
% mse, desired distance, the distance between each agent and its neighbors, close to the desired or not? 

\section{Cause of Failures}
\vspace{-3mm}
% \begin{figure}[ht]
%     \centering
%     \includegraphics[width=0.8\columnwidth]{templates/figs/flocking_circle/flocking_circle_error_plot.pdf}
%     \caption{Mean absolute error (MAE) of ten tests for a flock of 5 agents form a flock with circular pattern. The red dashed line shows the desired MAE.}
%     \label{fig:flock-circle-err}
% \end{figure}

%%%%%%%%%%%%% Traj only plot %%%%%%%%%%%%%%%%%%
\begin{figure}[bh!]
\vspace{-5mm}
    \centering
    
    \vspace{-3mm}
    \subfigure[Round 0, circle]{\includegraphics[width=0.24\columnwidth]{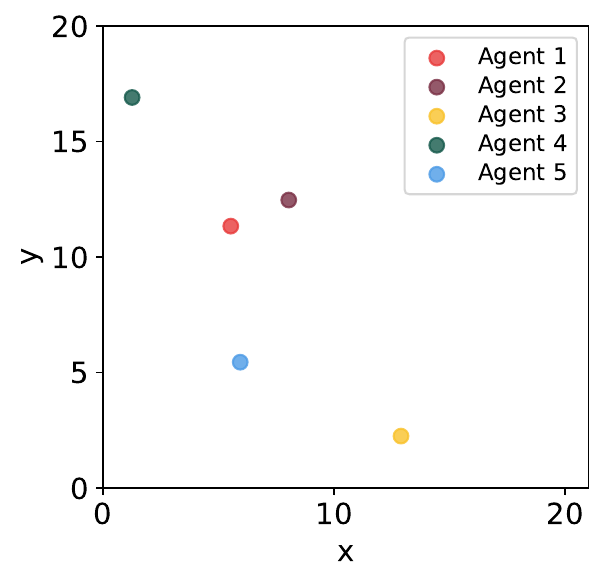}}
    \subfigure[Round 5, circle]{\includegraphics[width=0.24\columnwidth]{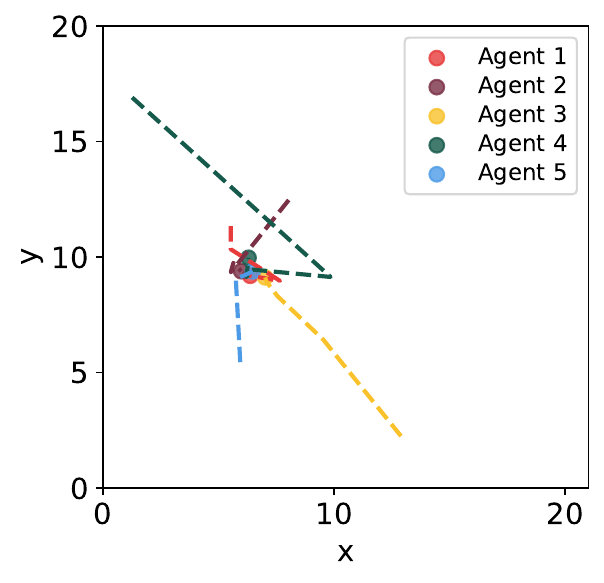}}
    \subfigure[Round 15, circle]{\includegraphics[width=0.24\columnwidth]{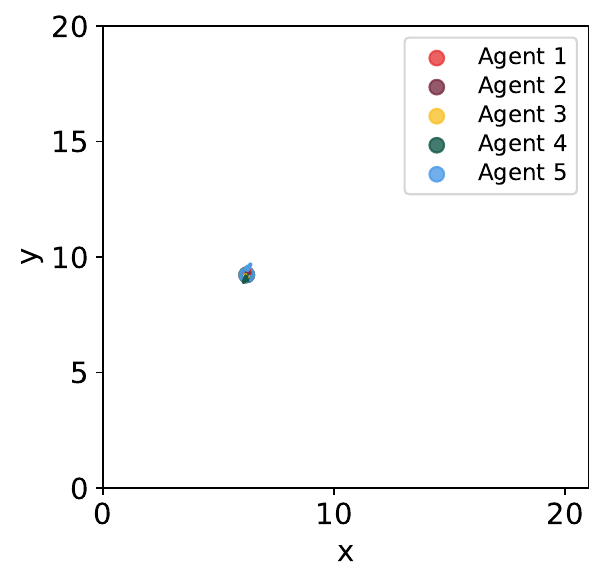}}
    \subfigure[Round 25, circle]{\includegraphics[width=0.24\columnwidth]{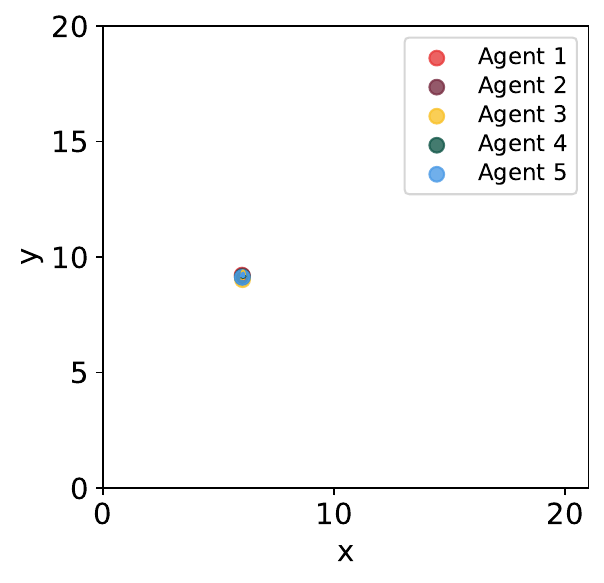}}
    % \\
    \vspace{-2mm}
    \subfigure[Round 0, $\alpha$-lattice]{\includegraphics[width=0.24\columnwidth]{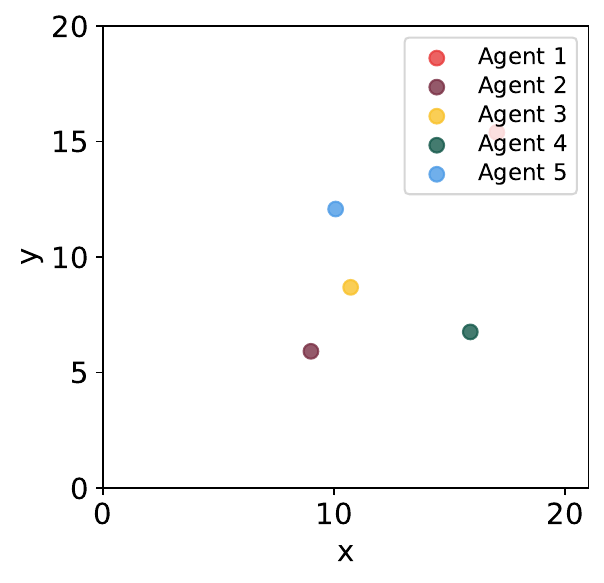}}
    \subfigure[Round 5, $\alpha$-lattice]{\includegraphics[width=0.24\columnwidth]{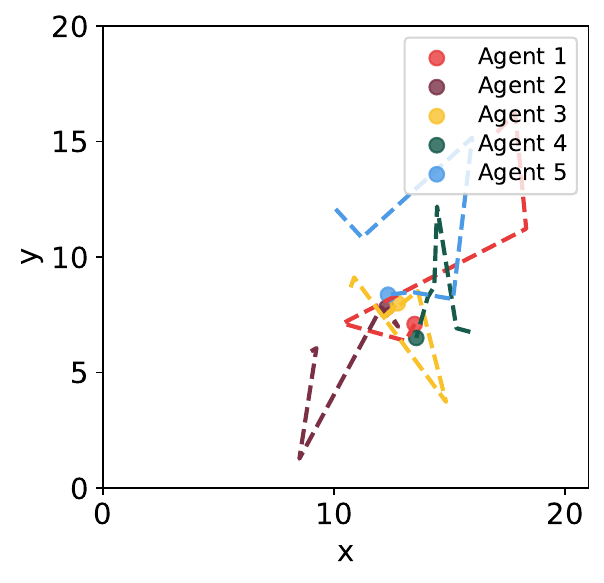}}
    \subfigure[Round 15, $\alpha$-lattice]{\includegraphics[width=0.24\columnwidth]{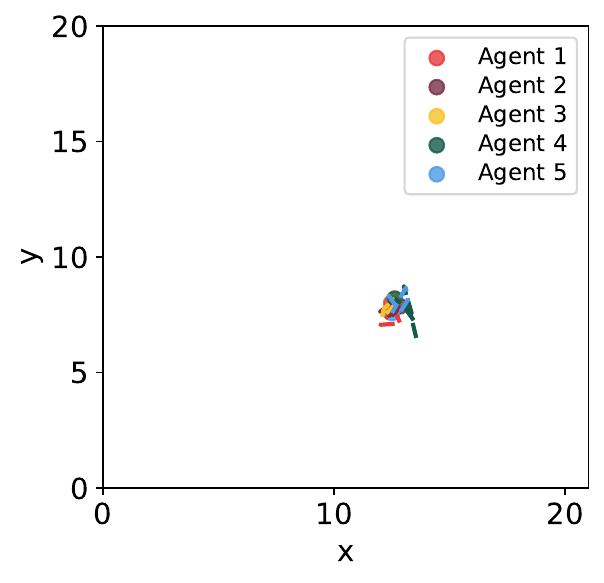}}
    \subfigure[Round 25, $\alpha$-lattice]{\includegraphics[width=0.24\columnwidth]{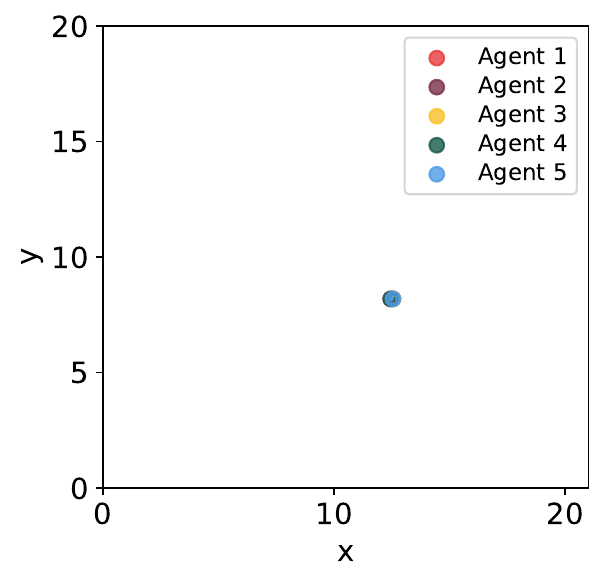}}

    \vspace{-2mm}
    \subfigure[Round 0, V-shape]{\includegraphics[width=0.24\columnwidth]{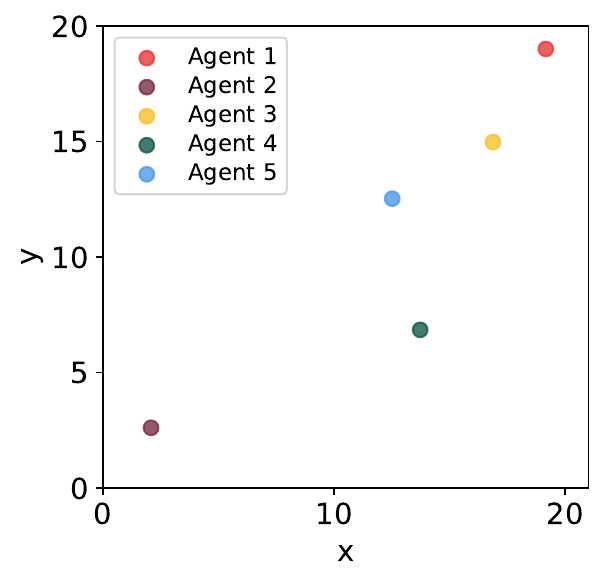}}
    \subfigure[Round 5, V-shape]{\includegraphics[width=0.24\columnwidth]{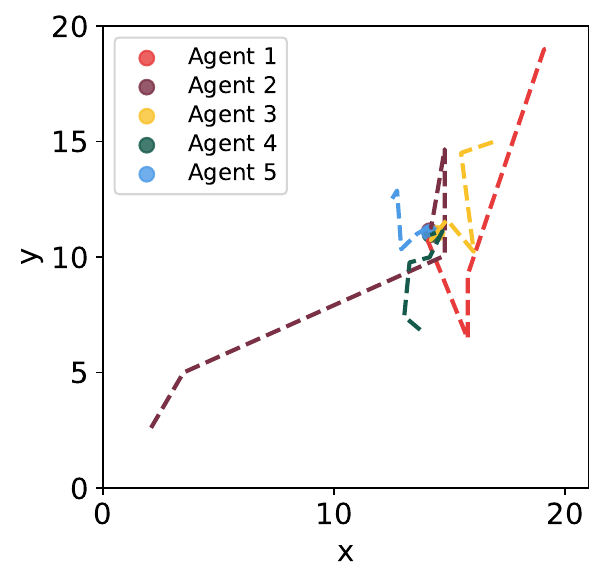}}
    \subfigure[Round 15, V-shape]{\includegraphics[width=0.24\columnwidth]{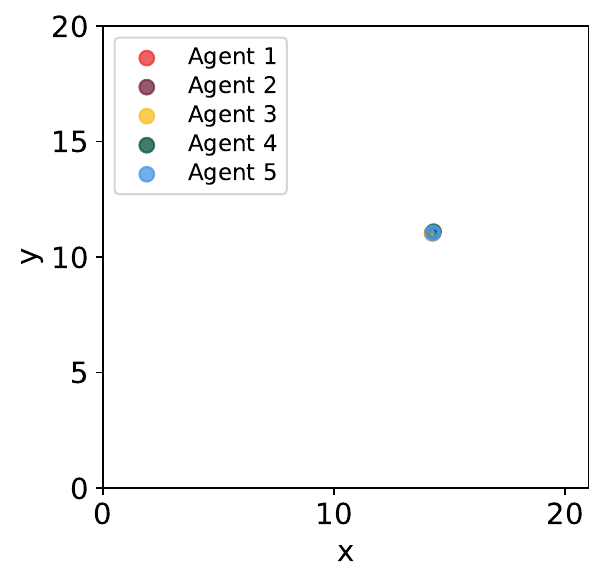}}
    \subfigure[Round 25, V-shape]{\includegraphics[width=0.24\columnwidth]{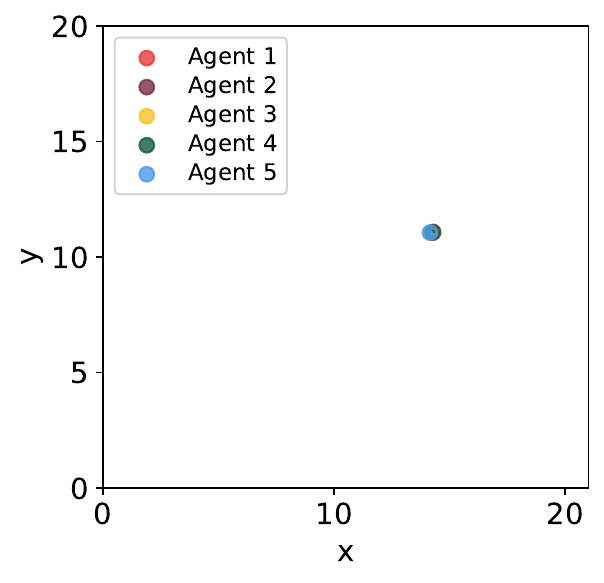}}

    \vspace{-3mm}
    \caption{Snapshots of the trajectory of selected tests consisting of five agents at different rounds. (a)-(d) represents the flock tasked to form a circle pattern. (e) - (h) represents the flock tasked to form an $\alpha$-lattice pattern. (i)-(l) represents the flock tasked to form a V-shape.}
    \label{fig:flocking_circle_traj}
\end{figure}
 All the tests use the GPT-3.5-Turbo-0613 model in this paper. We perform extensive tests on the more advanced model GPT-4-Turbo, but we have a high failure rate of around 80$\%$. The GPT-4-Turbo model will give more explanations, but it often ignores the output format we are given, resulting in failure to record the results. On the contrary, GPT-3.5-Turbo has a very low failure rate of around 5$\%$. The code implementation\footnote{\url{https://github.com/Zhourobotics/llms-for-flocking-pub}} and video of tests\footnote{\url{https://youtu.be/Oqa_F1TSitc}} are available online.
 % The code implementation and the video of the experiments is available on our GitHub page\footnote{\url{https://github.com/Zhourobotics/llms-for-flocking/}} .}
 % \vspace{-3mm}
 We first test multi-agent flocking with a group size of five to form a specific pattern. Each agent starts from a random position within the world and has its own LLM to decide the motion. The prompt contains the current positions of the agent and its neighbors, as well as the flock pattern the multi-agent group needs to form and the safe distance each agent must maintain. The LLM will make decisions based on the limited information provided and generate the new position the agent should go to. We specify different flock patterns, including circle, $\alpha$-lattice as described in~\cite{olfati2006flocking}, and V-shape. We perform ten tests on each flock pattern. We use mean absolute error (MAE) to measure the performance of the test. The mean absolute error is calculated by
\vspace{-3mm}
\begin{equation}    \label{eq:mae}
    \text{MAE} = \frac{1}{n}\sum^n_{i=1}\abs{\norm{p_i-p_{j}}-\epsilon}, ~i,j \in N, i\neq j.
\end{equation}
% \vspace{-3mm}
Here, $\norm{p_i - p_j}$ is the Euclidean distance between agent $i$ and $j$, where agent $j$ is the closest agent from agent $i$. $\epsilon$ is the desired distance between the closest agent. A lower MAE shows better cohesion and separation performances as the agents stay apart from the neighbors to avoid collision and are compact enough to satisfy the desired distance requirements. Since we are only controlling the positions of the agents, the velocity alignment performance will not be evaluated with metrics. %\LZ{we need to mention in the approach section why we don't do velocity.}

% \begin{figure}[tb]
%     \centering
%     \includegraphics[width=0.8\columnwidth]{templates/figs/triangle/triangle_0_static_error_plot.pdf}
%     \caption{MAE of ten tests for a flock of 3 active agents form a triangle pattern. The red dashed line shows the desired MAE.}
%     \label{fig:tri-0-err}
% \end{figure}

% \begin{figure*}[tb]
%     \centering
%     \subfigure[Five active agents circle]{\includegraphics[width=0.49\columnwidth]{templates/figs/flocking_circle/flocking_circle_error_plot.pdf} \label{fig:flock-circle-err}} 
%     \subfigure[Three active agents triangle]{\includegraphics[width=0.49\columnwidth]{templates/figs/triangle/triangle_0_static_error_plot.pdf} \label{fig:tri-0-err}} \\
%     \subfigure[Two active agents]{\includegraphics[width=0.49\columnwidth]{templates/figs/keep_dist/keep_distance_0_static_error_plot.pdf} \label{fig:dist-0-err}} 
%     \subfigure[One active agent, one stationary agent]{\includegraphics[width=0.49\columnwidth]{templates/figs/keep_dist/keep_distance_1_static_error_plot.pdf} \label{fig:dist-1-err}}
%     \caption{MAE of ten tests for a flock of 3 active agents form a triangle pattern. The red dashed line shows the desired MAE.}
    
% \end{figure*}

%%%%%%%%%%%%%%%% Traj w/ reasoning - Five Flock %%%%%%%%%%%%%%%%%
\begin{figure*}[th!] \label{fig:flock-trajectory}
\vspace{-2mm}
    \centering
    \subfigure[Round 5]{\includegraphics[width=0.85\columnwidth, trim={1cm 4cm 0.6cm 5.6cm},clip]{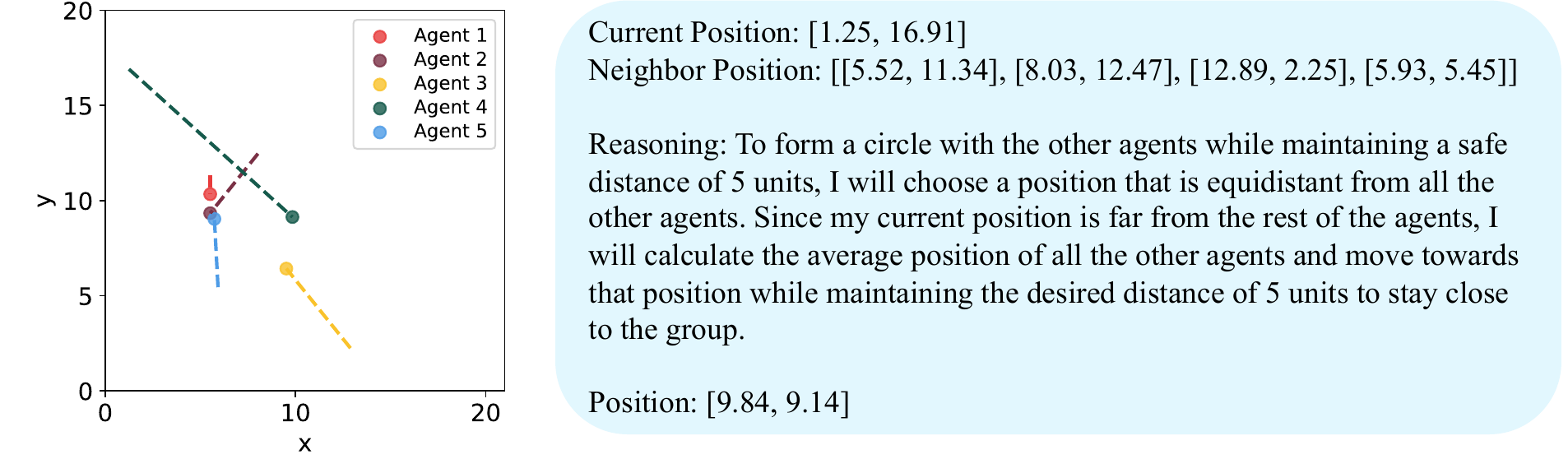} \label{subfig:flock-traj-1}} \\

    \vspace{-3mm}
    \subfigure[Round 6]{\includegraphics[width=0.85\columnwidth, trim={1cm 4cm 0.6cm 5.6cm},clip]{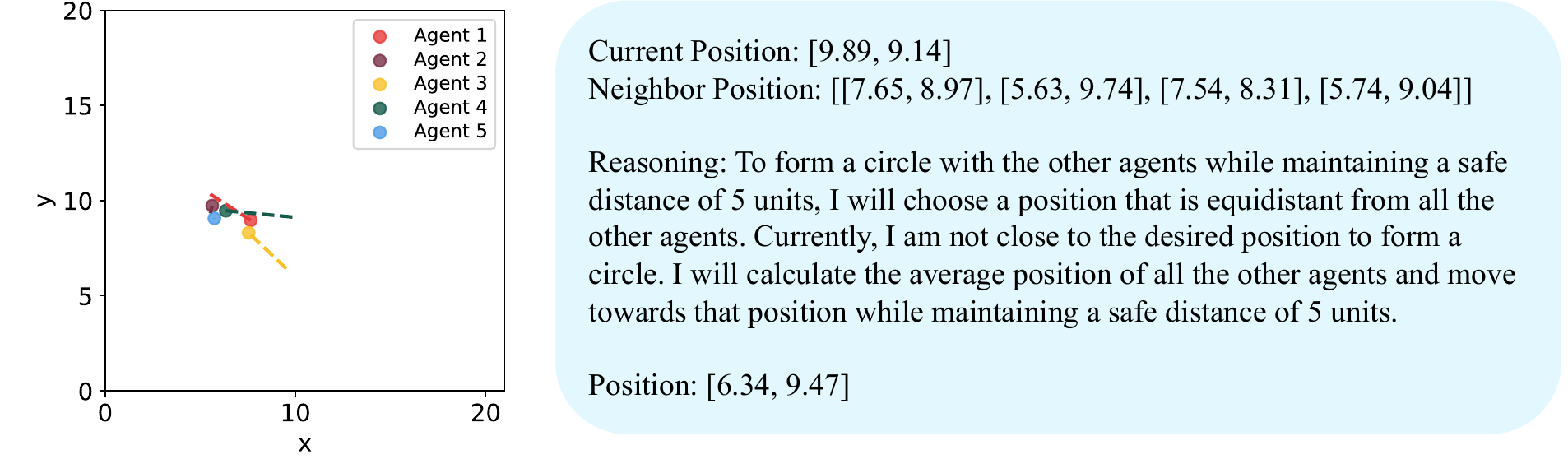} \label{subfig:flock-traj-2}} \\
    
    \vspace{-3mm}
    \subfigure[Round 7]{\includegraphics[width=0.85\columnwidth, trim={1cm 4cm 0.6cm 5.6cm},clip]{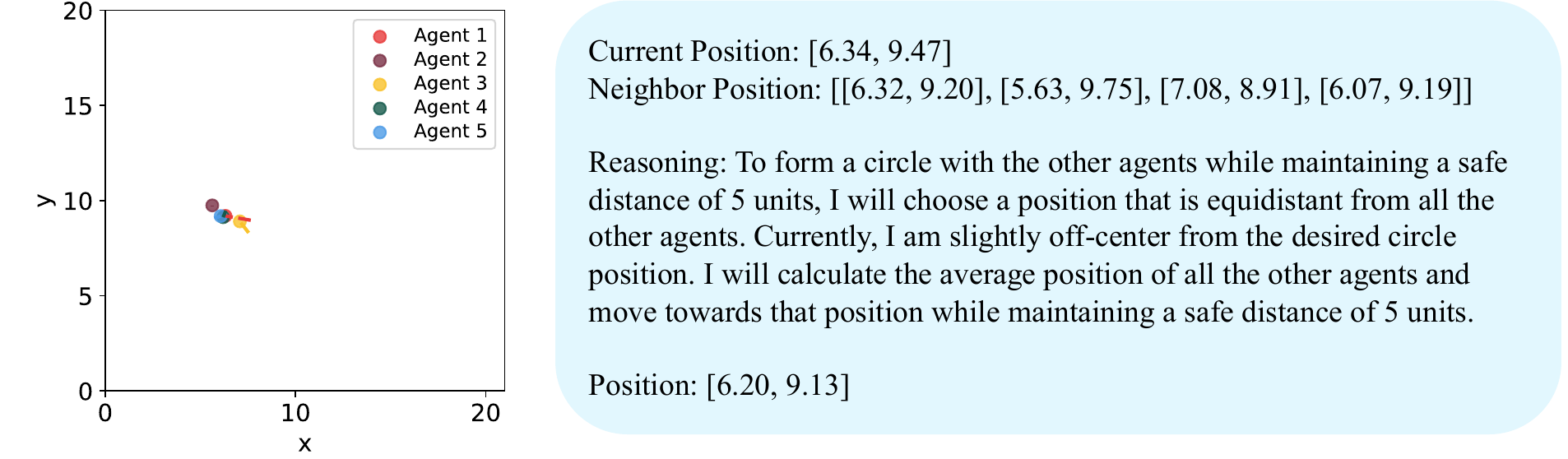} \label{subfig:flock-traj-3}} \\

    \vspace{-3mm}
    \subfigure[Round 8]{\includegraphics[width=0.85\columnwidth, trim={1cm 4cm 0.6cm 5.6cm},clip]{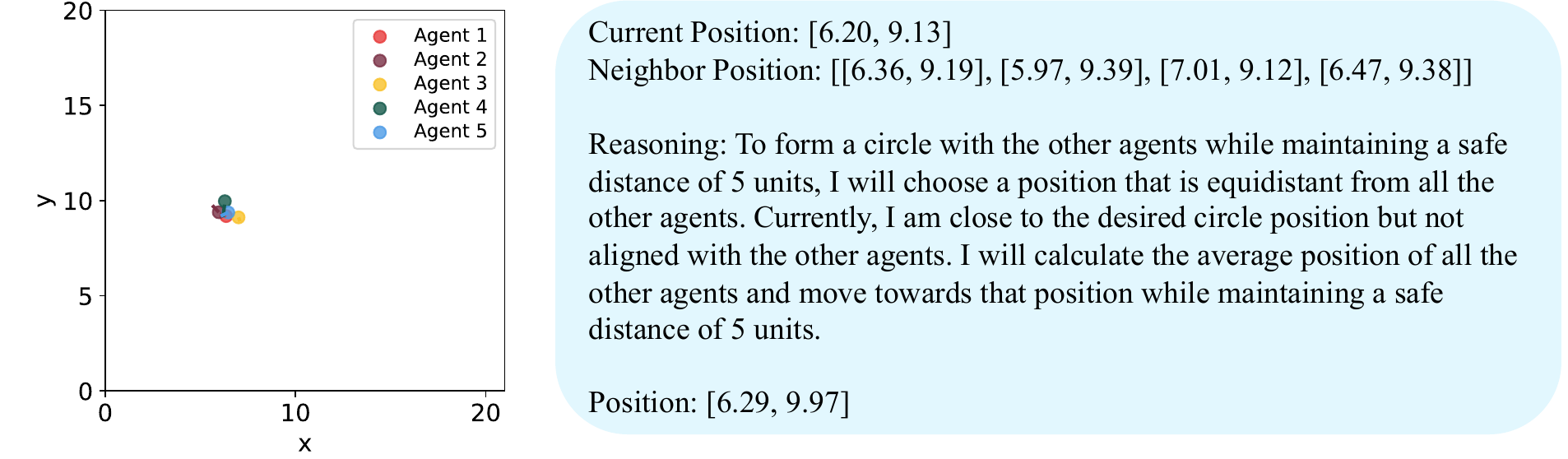}\label{subfig:flock-traj-4}} 
    % \subfigure[Round 5]{\includegraphics[width=0.7\columnwidth, trim={1cm 4cm 0.6cm 5.6cm},clip]{templates/figs/flocking_circle/flock_5.pdf}
    %\label{subfig:flock-traj-5}
    \vspace{-3mm}
    \caption{Snapshots of the trajectory from a selected test with five agents forming a circle. The desired distance between each agent is 5 units. Here, we focus on the reasoning and decisions of Agent 4's LLM.}
    \vspace{-7.5mm}
\end{figure*}

\noindent \textbf{Five-agent Tests.} Fig.~\ref{fig:flock-circle-err} shows the MAE for multi-agent flocking with a flocking pattern as a circle. The ideal perfect test will have MAE tends to be zero since all the agents are keeping away from the neighbors based on the desired distance described in the prompt. However, nearly 60$\%$ of the tests fail and gather to the same point instead of spread out to maintain the desired distance. Due to the page limit and the similar behavior, we only select the trajectories from three tests as shown in Fig.~\ref{fig:flocking_circle_traj}. The three rows represent selected tests forming a circle, $\alpha$-lattice, and V-shape patterns, respectively. The agents gather together to perform a consensus-like behavior instead of staying at the desired distance, which we consider a failure. We run ten tests for each pattern, but only one test is shown here due to the page limit and the similar behavior they have.

We then show a more detailed trajectory and the corresponding reasoning in Fig.3 for the same circle pattern test in Fig.~\ref{fig:flock-circle-err} and focus only on Agent 4's LLM as the reasoning of the other agents' LLMs is similar. In this test, we tell the agents to form a circle and keep a desired distance of 5 units from the closest agent. Intuitively, we would place the agents on the perimeter of the circle equidistant from the two closest neighbors. From Fig.~\ref{subfig:flock-traj-1}, we can observe Agent 4 choose to move closer to the rest of the agent. LLM takes the command of forming a circle as moving to an equidistant position from all other agents, which represents the center of the circle instead of some positions on the perimeter of the circle. This misunderstanding drives Agent 4 to perform consensus instead of flocking as shown in Fig.~\ref{subfig:flock-traj-2} to Fig.~\ref{subfig:flock-traj-4}. 

%%%%%%%%%%%%%%%% Traj w/ reason - Two agent %%%%%%%%%%%%%%
\begin{figure*}[!th] \label{fig:dist-traj}
\vspace{-3mm}
    \centering
    \subfigure[Round 5]{\includegraphics[width=0.85\columnwidth, trim={1cm 4cm 0.6cm 5.3cm},clip]{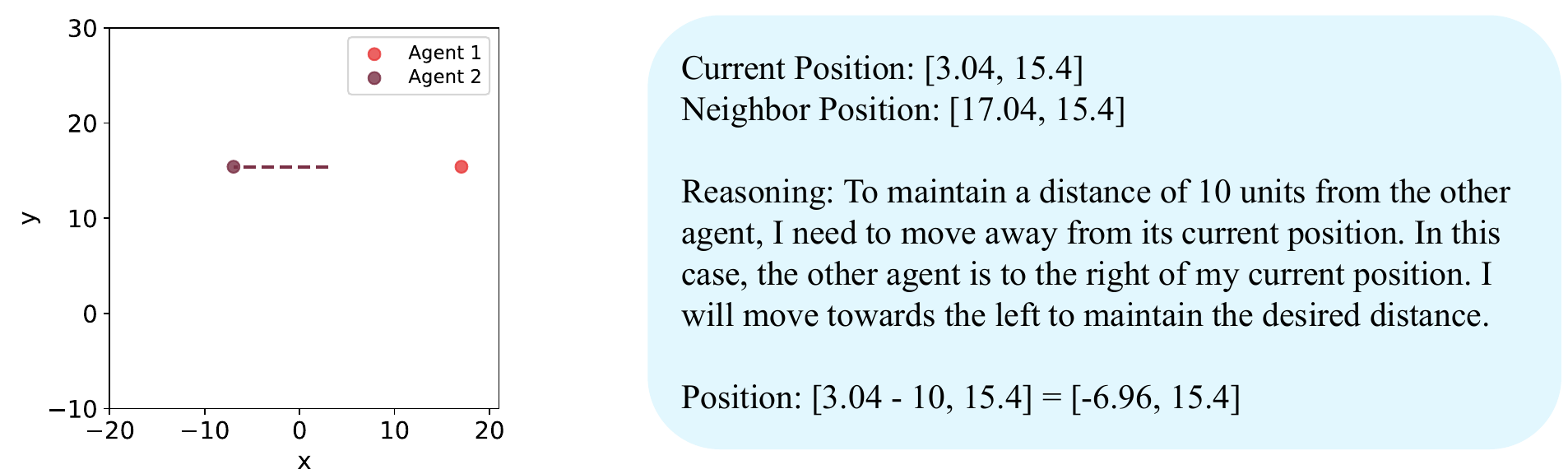} \label{subfig:dist-traj-1}} \\
    \vspace{-3mm}
    \subfigure[Round 6]{\includegraphics[width=0.85\columnwidth, trim={1cm 4cm 0.6cm 5.3cm},clip]{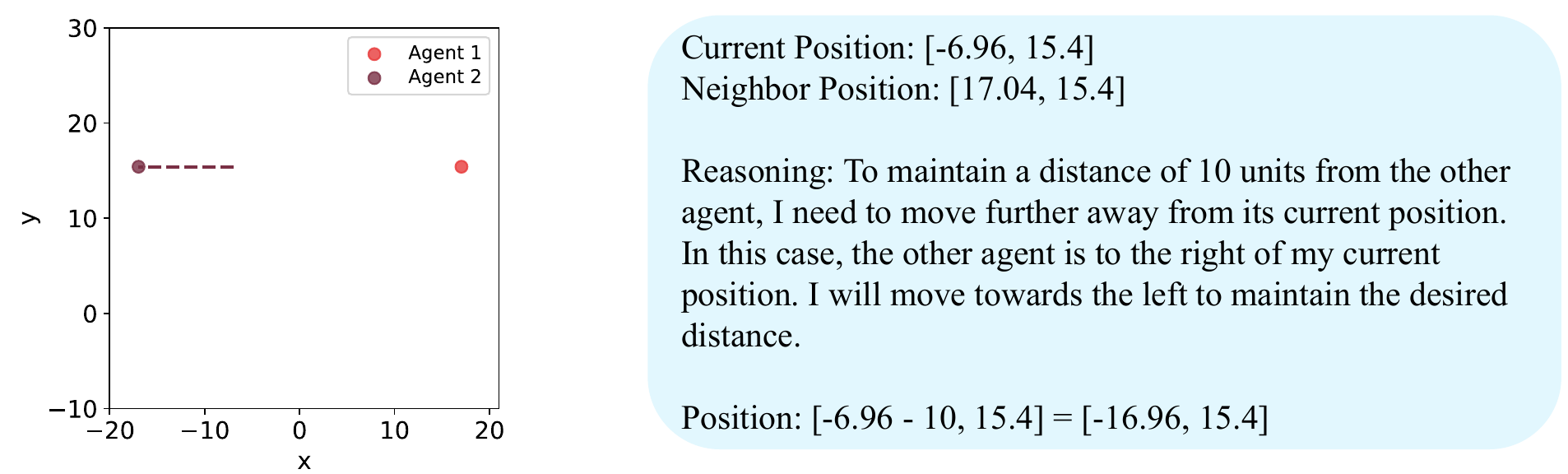} \label{subfig:dist-traj-2}} \\
    \vspace{-3mm}
    \subfigure[Round 7]{\includegraphics[width=0.85\columnwidth, trim={1cm 4cm 0.6cm 5.3cm},clip]{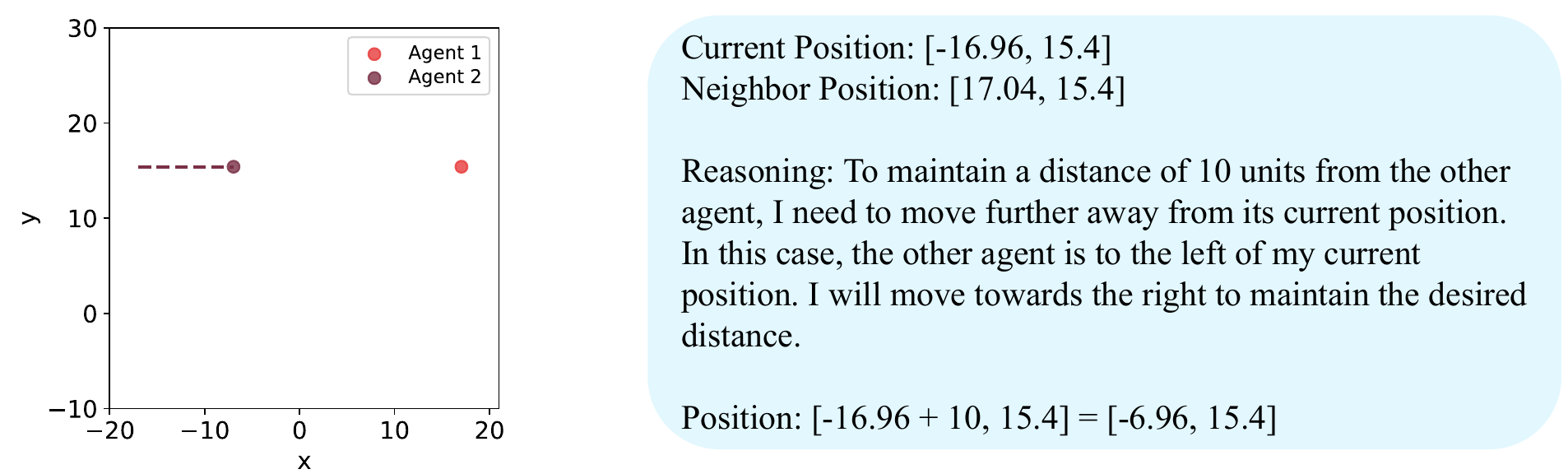} \label{subfig:dist-traj-3}} \\
    \vspace{-3mm}
    \subfigure[Round 8]{\includegraphics[width=0.85\columnwidth, trim={1cm 4cm 0.6cm 5.3cm},clip]{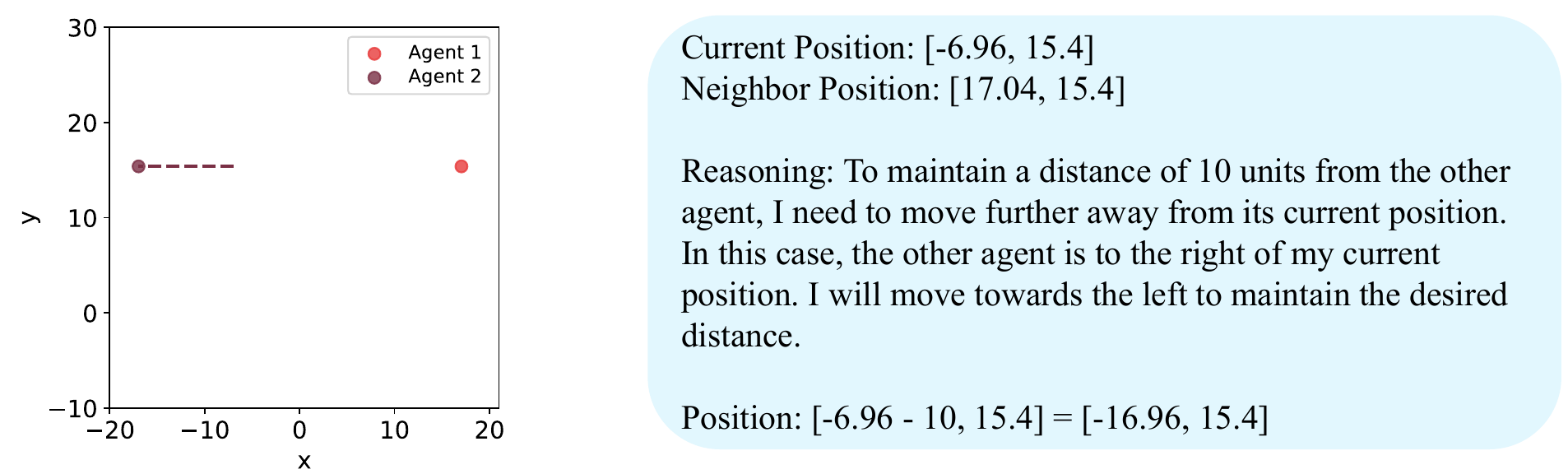} \label{subfig:dist-traj-4}} 
    % \\
    % \subfigure[Round 9]{\includegraphics[width=0.7\columnwidth, trim={1cm 4cm 0.6cm 5.3cm},clip]{templates/figs/keep_dist/dist_5.pdf} \label{subfig:dist-traj-5}}
    \vspace{-3mm}

    \caption{Snapshots of the trajectory from a sample test with one active and one stationary agent. The desired distance is 10 units. The corresponding reasoning from the active agent for each round is shown on the right.}
    \vspace{-5mm}
\end{figure*}

%%%%%%% Three agents -- Triangle %%%%%%%%%%%%%
\noindent \textbf{Three-agent Tests.} After the failed attempt to form a flock with five agents, we set up a simpler scenario, where three agents are staying at the 5 units' desired distance between neighbors. Our expectation for this scenario is the three agents will stay close together and form a triangle with an equal distance between each agent. The quantitative result with ten tests is shown in Fig.~\ref{fig:tri-0-err}, showing the MAE does not decrease to the desired value. The trajectory of the agents is similar to the flock of five agents, in which agents gather together instead of maintaining the desired distance. Different from the five-agent flocking scenario, flocking with three agents does not require the LLM to process complicated shape information and coordination. The triangle pattern is naturally formed when an agent maintains a desired distance from the other two neighbor agents. We then question if the LLM is capable of reasoning the distance and making corresponding decisions, which is crucial for flock formation.
\\
%%%%%%% Two agents -- Distance keeping %%%%%%%
\noindent \textbf{Two-agent Tests.} To test the ability to reason and understand distance, we set up the tests with two agents starting from random locations and trying to keep a desired distance between them. The MAE for the ten tests is shown in Fig.~\ref{fig:dist-0-err}. For all the tests, the desired distance is 10 units between two agents. Unlike the previous tests, where the agents gather to form a consensus, the two agents keeping a certain distance scenario is more random and spontaneous. Some tests show the consensus behavior, but other tests show the agents moving away from each other instead of staying at a certain distance. To further simplify the problem, we make one of the agents stationary, which means only one agent is moving to keep the desired distance. The MAE of ten tests is shown in Fig.~\ref{fig:dist-1-err}. A similar behavior is observed here in that the moving (or active) agent. Instead of keeping the desired distance, drives further away from the stationary agent.

% Peihan and Vishnu by March 26 

% \begin{itemize}
%     \item  try multi-agent first, 5 agents; GPT 4 and 3.5; circle; lattice, V-shape, line. 
%     \item try three agents; GPT 4 and 3.5; triangle (two are stationary and 1 is moving), line
%     \item two agents; GPT 4 and 3.5; two moving; one stationary one moving
    
%     \item Peihan, Vishnu, Bhavana
% \end{itemize}  

% \vspace{-1mm}
\noindent \textbf{Reasoning Capability.}
% Peihan, Vishnu, Bhavana, Dan
% can't reason about distance, desired shape, etc. 
From all the test scenarios above, we think the LLMs are limited in reasoning the given position and distance information and making corresponding decisions. Fig.~\ref{fig:dist-traj} shows the trajectory snapshot of one test with one stationary agent and one active agent as well as the corresponding reasoning provided by the active agent explaining the decision. Fig.~\ref{subfig:dist-traj-1} shows that in Round 6 of the test, the active agent has already moved away from the stationary agent. The position of the stationary agent is [17.04, 15.4], and the current position of the active agent is [-16.96, 15.4]. The distance between two agents is $d=17.04-(-16.96)=34$ units, which exceeds the desired distance of 10 units. However, given the positions of both agents, the LLM of the active agent falsely reasons the situation and thinks the stationary agent is to the left of the active agent. With the false reasoning, LLM makes a decision to move the active agent to the right to drive further from the stationary agent. Fig.~\ref{subfig:dist-traj-2} to \ref{subfig:dist-traj-4} show the LLM continues to have false situation awareness and makes the wrong decision to switch position back and forth. One possible explanation is the LLM cannot reason the spatial positions giving only the coordinates of the points. From the example we give, the LLM probably does not reason the negative sign in the $x$-coordinate of the active agent and makes a decision based on a false understanding. Chen \etal \cite{chen2024solving} discussed that LLMs generally have an unsatisfactory performance in reasoning lists of tuples containing coordinates. This lack of reasoning ability is the main resistance for the LLM to effectively solve the flocking problem. Currently, solving the multi-agent flocking problem by using multiple LLMs as individual decision-makers with the GPT-3.5-Turbo model seems infeasible. But the LLMs is a rapidly evolving field, and a model with stronger reasoning capability might have a better performance on this specific task.

% \subsection{Various Sizes of Context} 
% Peihan, Vishnu, Bhavana

\vspace{-3mm}

\section{Conclusion and Future Work}
\vspace{-3mm}
This work investigated the challenges faced by LLMs in solving the multi-agent flocking problem. It demonstrates that generic LLMs are unable to solve multi-agent flocking or even its component problems such as coordinating movements, maintaining a shape, and keeping a distance through extensive experiments. We have to conclude that the current LLMs (e.g., GPT-3.5-Turbo) that are not fine-tuned do not yet possess a good understanding of spatial and collaborative reasoning. As a result, they cannot be used to make decentralized decisions for agents to achieve flocking behaviors. However, we hypothesize that an LLM with better capabilities in these areas could potentially overcome these challenges. Solving this could not only make LLMs more generally intelligent but also be applied to solve more complex multi-agent problems.

% Thus focus should be directed at improving the spatial awareness of LLMs as well as their reasoning capacity.

% Peihan: April 1
% Vishnu: April 2
% Lifeng: April 2. 
To address LLMs' limitations in multi-agent flocking demonstrated by failures in spatial reasoning, future work should focus on developing more sophisticated LLMs, fine-tuning existing models, and possibly integrating visual data for enhanced spatial understanding and reasoning capability. This paper also shares many similar challenges identified in another study exploring the challenges of LLM-based pathfinding tasks~\cite{chen2024solving}. Further research could explore breaking down complex tasks into simpler components LLMs can more readily solve, enhancing context management, spatial awareness, and strategic planning capabilities in a decentralized manner. 

%
% ---- Bibliography ----
%
% \PL{Bibtext}
\bibliographystyle{plain}
\bibliography{author}

\end{document}